\documentclass[conference]{IEEEtran}
\IEEEoverridecommandlockouts
\usepackage{cite}
\usepackage{amsmath,amssymb,amsfonts}
\usepackage{algorithmic}
\usepackage[]{graphicx}
\graphicspath{ {./figs/} }
\usepackage{textcomp}
\usepackage{xcolor}

\usepackage{comment}

\usepackage{epsfig}
\usepackage{subcaption}
\usepackage{url}
\usepackage{array}
\usepackage{tabularx}
\usepackage{hhline}
\usepackage{xspace}
\usepackage{upgreek}
\usepackage{makecell}
\usepackage{setspace}

\newcounter{numberlistc}

\newenvironment{numberlist}
    {   \setcounter{numberlistc}{0}
        \begin{list}{\arabic{numberlistc}.}
        {\usecounter{numberlistc}
        \setlength{\parsep}{0pt}
        \setlength{\topsep}{3pt}
        \setlength{\itemsep}{0pt}}
        }{ \end{list} }
\newcounter{itemlistc}

\begin{document}
%
\title{EM-GAN: Fast Stress Analysis for Multi-Segment Interconnect Using Generative Adversarial Networks}


\author{Wentian Jin$^{1}$, Sheriff Sadiqbatcha$^{1}$, Jinwei
Zhang$^{1}$ and Sheldon X.-D. Tan$^{1}$
\thanks{$^{1}$Wentian Jin, Sheriff Sadiqbatcha, Jinwei Zhang and
Sheldon X.-D. Tan are with the Department of Electrical and Computer
Engineering, University of California, Riverside, CA 92521.}%
}

\maketitle

\begin{abstract}
  In this paper, we propose a fast transient hydrostatic stress
  analysis for electromigration (EM) failure assessment for
  multi-segment interconnects using generative adversarial networks
  (GANs). Our work leverages the image synthesis feature of GAN-based
  generative deep neural networks. The stress evaluation of
  multi-segment interconnects, modeled by partial differential
  equations, can be viewed as time-varying 2D-images-to-image problem
  where the input is the multi-segment interconnects topology with
  current densities and the output is the EM stress distribution in
  those wire segments at the given aging time.  Based on this
  observation, we train conditional GAN model using the images of many
  self-generated multi-segment wires and wire current densities and
  aging time (as conditions) against the COMSOL simulation results.
  Different hyperparameters of GAN were studied and compared. The
  proposed algorithm, called {\it EM-GAN}, can quickly give accurate
  stress distribution of a general multi-segment wire tree for a given
  aging time, which is important for full-chip fast EM failure
  assessment.  Our experimental results show that the EM-GAN shows
  6.6\% averaged error compared to COMSOL simulation results with
  orders of magnitude speedup. It also delivers $8.3 \times$ speedup
  over state-of-the-art analytic based EM analysis solver.

\end{abstract}

\section{Introduction}
\label{sec:intro}

Electromigration (EM)  is a primary long-term reliability concern for
copper-based back-end-of-the-line interconnects used in modern
semiconductor chips. As predicted by ITRS, EM is projected to only get
worse in future technology nodes~\cite{ITRS'15}.  This, as with many
other reliability effects, is due to the continued trend of
feature-size reduction and rapid integration which ultimately affects
the critical sizes for the EM failure process. EM-related aging and
reliability will become worse for current 7nm and below technologies.
As a result, it is crucial to ensure the reliability of the VLSI chips
during their projected lifetimes.



Due to its growing importance, considerable recent research has
focused on fast EM analysis techniques.  It is well accepted that
existing Black and Blech-based EM
models~\cite{Black:1969fc,Blech:1976ko} are overly conservative and
can only work for single wire
segment~\cite{Hauschildt:2013cv,Sukharev:2013tq}. Recently, a number
of physics-based EM model and analysis techniques have been
proposed~\cite{deOrio:2010,HuangTan:TCAD'16,sukharev2016postvoiding,ChenTan:TCAD'16,MishraSapatnekar:2016DAC,ChenTan:TDMR'17,Chatterjee:2018TCAD,CookSun:TVSI'18,WangSun:ICCAD'17,
ZhaoTan:TVLSI'18,
Abbasinasab:DAC'2018,ChenTan:TVLSI'19,TanTahoori:Book'19}. At the
center of those methods is to solve partial differential equation
(called Korhonen's equation) of stress in the confined metal wire
segments in a general interconnect tree~\cite{Korhonen:jap1993}.
Although many numerical approaches such as finite
method~\cite{Chatterjee:2018TCAD,CookSun:TVSI'18}, finite element
methods~\cite{deOrio:2010,ZhaoTan:TVLSI'18} and analytic or
semi-analytic solutions~\cite{ChenTan:TCAD'16, ChenTan:TDMR'17,
WangSun:ICCAD'17, Abbasinasab:DAC'2018, ChenTan:TVLSI'19} were
proposed, these methods still suffer the high computing costs or can
only apply to some special cases, which hinder this applications for
full-chip EM validation and signoff analysis.

On the other hand, deep neural networks (DNN) have propelled an
evolution in machine learning fields and redefined many existing
applications with new human-level AI capabilities. DNNs such as
convolution neural networks (CNN) have recently been applied to many
cognitive applications such as visual object recognition, object
detection, speech recognition, natural language understanding, and
etc. due to dramatic accuracy improvements in those
tasks~\cite{LeCun:2015dt}. Recently, generative adversarial networks
(GAN)~\cite{Goodfellow:Book'2016} gained much traction as it can learn
features (latent representation) without extensively annotated
training data. The representations learned by GANs may be used in a
variety of applications, including image synthesis, semantic image
editing, style transfer, image super-resolution, and classification.
Recently GAN-based methods have been applied  for VLSI physical
designs such as for layout lithography analysis~\cite{Ye:DAC'2019} and
sub-resolution assist feature generation~\cite{Alawieh:DAC'2019}, for
analog layout well generation~\cite{Xu:DAC'2019} and for routing
congestion estimation~\cite{Yu:DAC'2019}.

In this work, we propose a fast transient hydrostatic stress analysis
technique for EM failure assessment of multi-segment interconnects
using GANs. The new contributions are as follows:
\begin{numberlist}
\item We propose a fast GAN-based stress analysis solver, called {\it
    EM-GAN} for multi-segment interconnect wire tree. We treat the
    partial differential equation solving process as a time-varying
    2D-image-to-image process where the input is the multi-segment
    interconnects topology with current densities and aging time and
    the output is the EM stress distribution in those wire segments at
    the given aging time.
  
\item We design the architecture and hyper parameters of the EM-GAN
  solver. Different hyper parameters of GAN were studied and compared.
  We use current densities of wire segment and aging time as the
  conditions for the conditional GAN. The resulting {\it EM-GAN} can
  quickly give accurate stress distribution of any multi-segment wires
  for a given aging time.

  \item Our experimental results show that the EM-GAN has 6.6\%
    averaged error compared to COMSOL~\cite{comsol} simulation results
    with orders of magnitude speedup. It also delivers $8.3 \times$
    speedup over recently proposed state-of-the-art analytic based EM
    analysis solver~\cite{ChenTan:TVLSI'19}.

\end{numberlist}

\section{Physics-based EM modeling and analysis}
\label{sec:em_modeling}

EM is the process of metal atoms migrating along the direction of the
applied electric field in confined metal interconnect wires due to the
momentum transfer between the conducting electrons and lattice atoms.
Under EM, the aforementioned momentum transfer leads to the buildup of
hydrostatic stress in the confined metal wires. When this stress
reaches a critical level, the aforementioned migration of atoms is
initiated. Over time, this migration leaves behind a depletion of
atoms (or void) at the cathode terminal of the wire and an
accumulation of atoms (or hillock) at the anode terminal. This
eventually leads to failure due to an open or short circuit
respectively.

\begin{figure}[!htb]
\centering
\includegraphics[width=0.45\columnwidth]{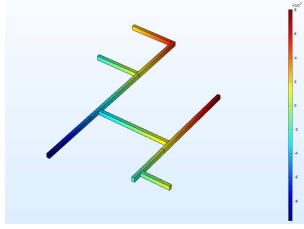}
\caption{Multi-segment wire with EM stress distribution}
\label{fig:multi_seg}
\end{figure}

Traditionally, the industry standard model to predicting the
time-to-failure (TTF) under EM are based on empirical or statistical
models, the most well known of which are Black's
equation~\cite{Black:1969fc} and Blech's limit~\cite{Blech:1976ko}.
However those models have been shown to be overly conservative,
applicable only to single wire segment, and therefore lead to
unnecessary over-design with large overheads~\cite{Sukharev:2013tq}.
To mitigate this problem, EM modeling starts with the first principles
of stress physics in the confined metal wires start to gain many
tractions~\cite{TanTahoori:Book'19}. Such physics-based EM modeling
analysis is centering around solving the partial differential equation
with blocked terminal boundary conditions for general multi-segment
interconnects as shown in Fig.~\ref{fig:multi_seg}.


Specificically, we assume that a general interconnect wire has $n$
nodes, including $p$ interior junction nodes
$x_r \in \{x_{r1}, x_{r2}, ..., x_{rp}\}$ and $q$ block terminals
$x_b \in \{x_{b1}, x_{b2}, ..., x_{bq}\}$. Then the Korhonen's
PDE~\cite{Korhonen:jap1993} for the nucleation phase can be written in
following multi-segment format: \begin{equation}
  \begin{aligned}
  & \frac{\partial \sigma_{ij}(x,t)}{\partial t}=\frac{\partial }{\partial x}\left[\kappa_{ij}(\frac{\partial
    \sigma_{ij}(x,t)}{\partial x} + G_{ij})\right] ,t>0; \\
& BC: \,\, \sigma_{ij_1}(x_i,t)=\sigma_{ij_2}(x_i,t),t>0; \\
& BC: \sum_{ij} w_{ij} \kappa_{ij}(\frac{\partial \sigma_{ij}(x,t)}{\partial x} \bigg |_{x=x_r}+G_{ij}) \cdot n_r = 0, t>0 \\
& BC: \kappa_{ij}(\frac{\partial \sigma_{ij}(x,t)}{\partial x} \bigg
|_{x=x_b}+G_{ij}) =  0, t>0; \\
& IC: \sigma_{ij}(x,0)= \sigma_{ij,T}
\label{eq:korhonen_tv_nz_initial_stress}
\end{aligned}
\end{equation}
where $\sigma(x,t)$ is the hydrostatic stress for branch $ij$ from
nodes $i$ and $j$, $n_r$ represents the unit inward normal direction
of the interior junction node $r$ on branch $ij$, the value of which
is $+1$ for right direction and $-1$ for left direction of branch with
assumption of $x_i < x_j$, $G = \frac{Eq*}{\Omega}$ is the EM driving
force, $w$ is the width of the branch, and $\kappa = D_a B\Omega/k_B
T$ is the diffusivity of stress. $E$ is the electric field, $q*$ is
the effective charge. $D_a = D_0\exp(\frac{-E_a}{k_B T})$, which is
the effective atomic diffusion coefficient. $D_0$ is the
pre-exponential factor, $B$ is the effective bulk elasticity modulus,
$\Omega$ is the atomic lattice volume, $k_B$ is the Boltzmann's
constant, $T$ is the absolute temperature, $E_a$ is the EM activation
energy. $\sigma_T$ is the initial thermal-induced residual stress in
each wire segment.

In general, numerical approaches such as finite difference, finite
element based approaches are required to solve the PDE in
\eqref{eq:korhonen_tv_nz_initial_stress}, which are expensive and time
consuming. The recently proposed semi-analytic solutions can still be
expensive as the eigenvalues have to be computed by numerical
approaches~\cite{WangSun:ICCAD'17, ChenTan:TVLSI'19}.

On the other hand, we can view the PDE solving process for a
multi-segment wire shown in Fig.~\ref{fig:multi_seg} as image
synthesis process, in which the deep neural networks (DNN) can
automatically extract features reflecting the physics-law of stress
evolution in the confined metal wire, then we can use the DNN network
to map the input images of interconnect wires with stressing current
or voltages to the stress distributions of wire segment for any given
aging time.  

\section{Data Preparation}
\label{sec:dataprep}

For machine learning based approaches, one crucial aspect is
sufficient training data. For our GAN-based EM stress estimation, it
also requires a large amount of interconnect topologies with various
current densities and corresponding ground truth EM stress
distribution data from which the model can learn the transformation
scheme in between. In what follows, we present the training data
required by our model and the method to collect them. Some necessary
pre-processing methods performed on the training set prior to feeding
them to the model will also be discussed.

\begin{figure}[!htb]
\vspace{-0.1in}
\centering
\begin{subfigure}{0.4\columnwidth}
\centering
\includegraphics[width=1\columnwidth]{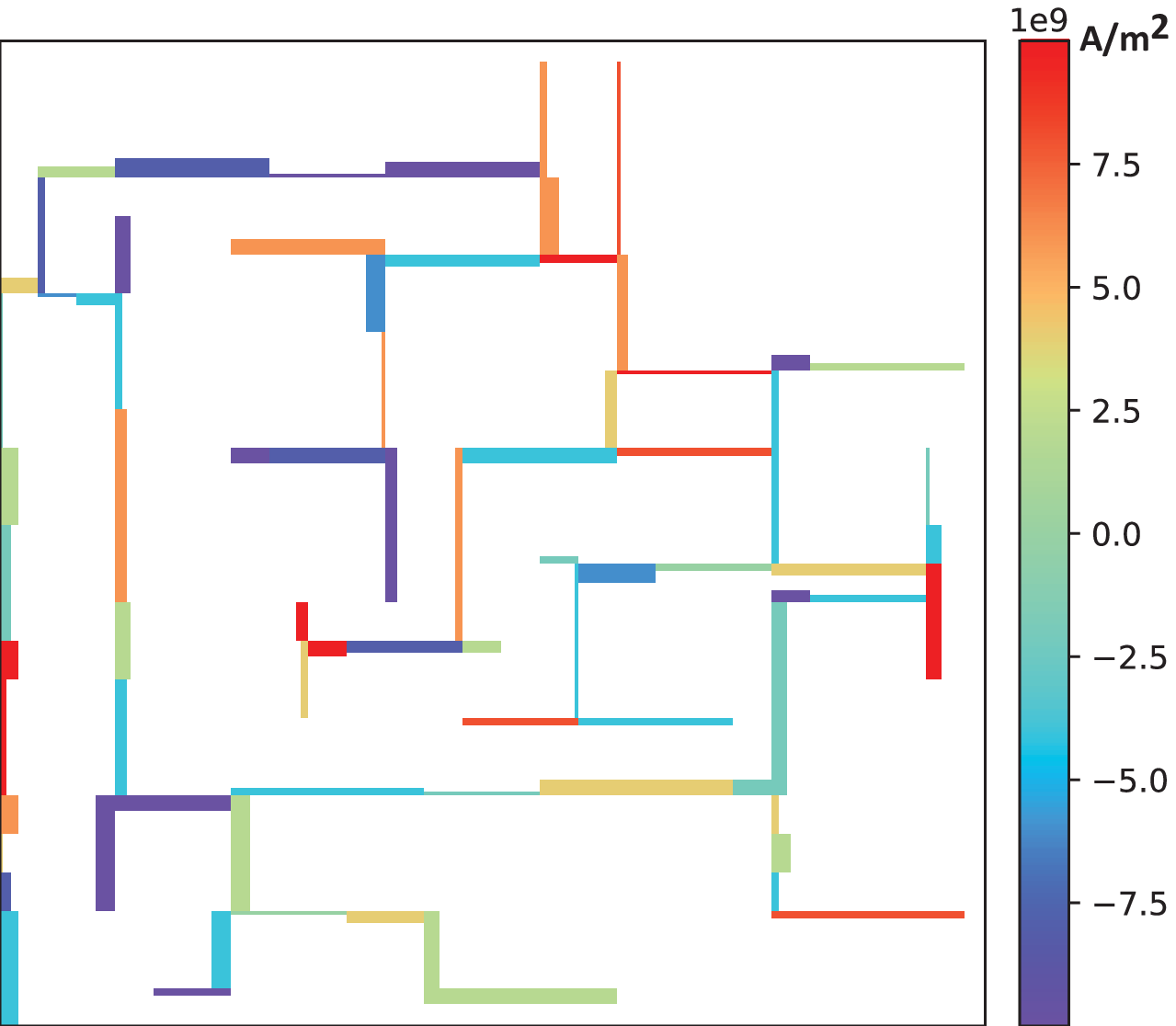}
\caption{}\label{fig:current}
\end{subfigure}
\hspace{+0.2in}
\begin{subfigure}{0.4\columnwidth}
\centering
\includegraphics[width=1\columnwidth]{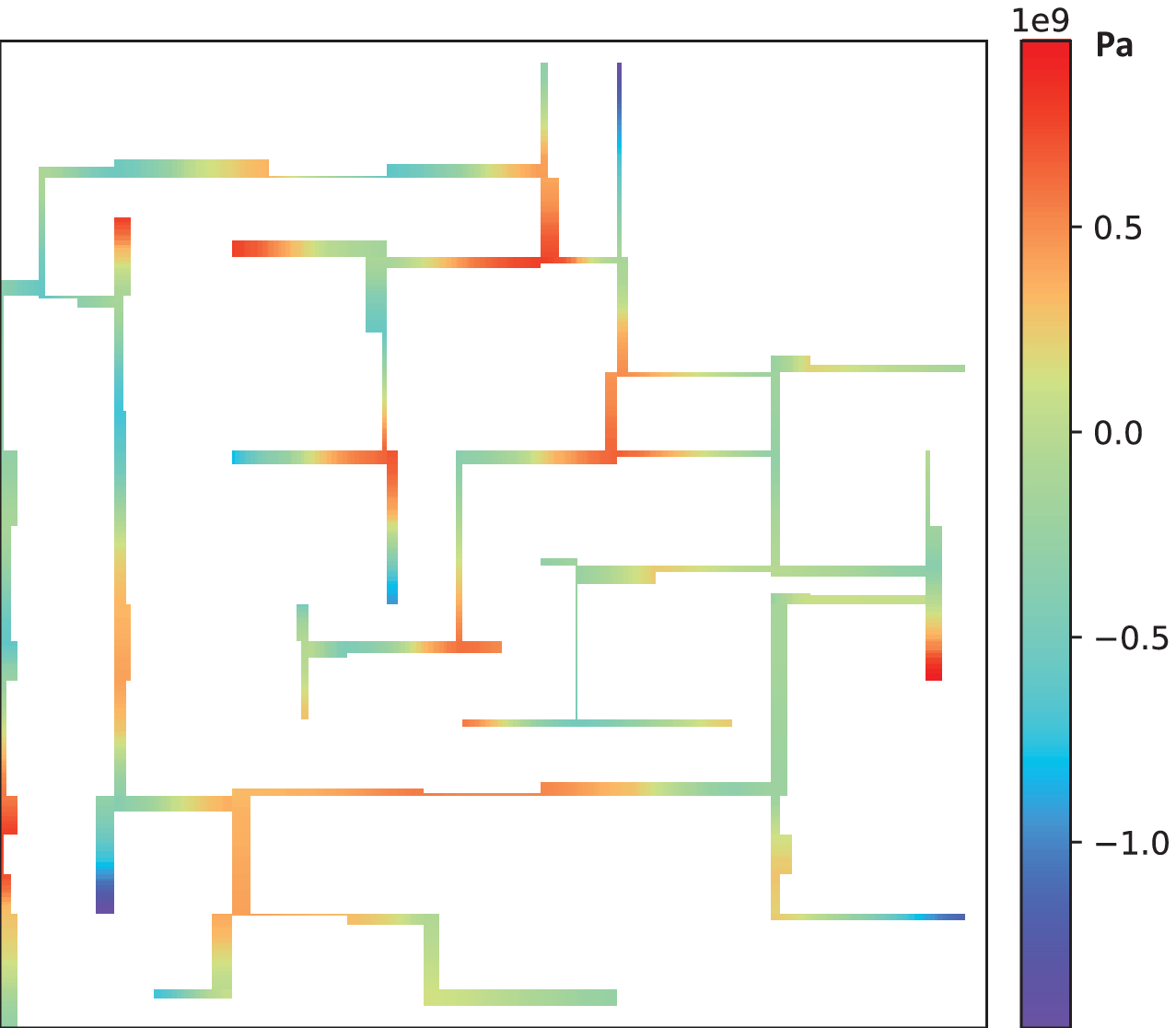}
\caption{}\label{fig:stress}
\end{subfigure}
\caption{Illustration of a random training data: (a) Interconnect
  topology with current density (b) EM stress distribution at 10th
  aging year.}
\label{fig:data_illustration}
\vspace{-0.05in}
\end{figure}

To achieve the abundance in the training set, we randomly generated
2500 different topologies of multi-segment interconnects with various
wire width, number of branches and current densities. The raw topology
and the current density data are separately stored in numerical
format. They are given to the COMSOL, which is a finite element
method(FEM) solver, as input and the resulting EM stress distributions
at 1 to 10 discrete aging years are saved as ground truth. As
mentioned in Section~\ref{sec:intro}, to leverage the GAN model, we
view this solving process as an image-to-image problem. Both
interconnects topology and current information can be synthesized into
a 2D-image (called {\it design} in this work) shown in
Fig.~\ref{fig:current}. The image actually has only one channel
instead of red-green-blue (RGB) channels and the color in
Fig.~\ref{fig:current} is only for illustration purpose. Every current
density value is filled into its position in the topology and zeroes
are padded to all positions without an interconnect. To make it easier
for a neural network to handle, we fixed the dimensions of each design
to $256 \times 256$ $\upmu$m$^{2}$ and make each pixel represent 1
$\upmu$m. Such setting does not restrict our work to only small
dimensions as in real applications, bulk interconnect system may be
divided into small pieces with partitioning algorithms for parallel
simulation. The ground truth EM stress distribution is also
synthesized into single-channel images with pixels filled with stress
values, as shown in Fig.~\ref{fig:stress}. Our training set contains
25000 samples where each sample is a (input design image, target EM
stress image) pair.

As the pixels in our images are not RGB colors but real current
density or EM stress values instead, they can range drastically from
magnitude of $-10^{9} $ to $10^{9} $ in which minus denotes the
current directions toward left and down sides. Such a large numerical
range is not suitable for neural networks and requires to be scaled
down. In this work, we rescaled all samples in the training set to
mean value of 0 and standard deviation of 1 using data standardization
method. It squeezes all data to the range of -7 to 7 and most of which
are around zero.




\section{CGAN-based Current Density to EM Stress Estimation}
\label{sec:cganbasedemstressestimation}

\subsection{GANs and CGANs}
\label{sec:gansandcgans}
The GAN is a neural network model that is used in unsupervised machine
learning tasks and was created by Ian
Goodfellow~\cite{Goodfellow:NIPS'14}. A traditional GAN is composed of
two separate deep neural networks, one is generator $\mathbf{G}$ and
the other is discriminator $\mathbf{D}$. $\mathbf{G}$ is trained to
generate ``real-like" output that resembles the data in the training
set while, on the contrary, $\mathbf{D}$ is trained as a judge to
distinguish between the real and generated data. The generator in a
conventional GAN takes random noise vector $\mathbf{z}$ as input and
transforms it into the output $\mathbf{G(z)}$. Both real and generated
data are alternatively given to the discriminator which is typically a
deep binary classifier and output a ``score''. The input data are
classified as ``real" or ``fake" based on the score which also serves
as part of the loss function. It is used to train both $\mathbf{G}$
and $\mathbf{D}$ through back propagation. The training processes of
both networks are performed simultaneously in an alternative way to
ensure neither of them is lagging too much behind the other until an
equilibrium is reached. 

The conditional GAN is a GAN working in the conditional setting and
learns a conditional generative model. Unlike the conventional GAN,
the input of the generator is a combination of both condition vector
$\mathbf{x}$ and random noise vector $\mathbf{z}$ and the output is
denoted as $\mathbf{G(x,z)}$. The main difference compared to GAN is
that both $\mathbf{G}$ and $\mathbf{D}$ in CGAN are conditioned on the
vector $\mathbf{x}$. CGAN has been shown working perfectly in
image-to-image translation works where the input image is seen as the
condition and constrains the image generation process. In our context,
the generated EM stress distribution is conditioned on the input
current density and given aging time according to physics-law of
stress evolution, which is highly suitable for the CGAN model.

The training of GAN is quite challenging given the fact that such
process is a minimax game between two separate neural networks. The
objective function of either generator or discriminator is influenced
not only by itself but also by its opponent. As a result, when the
discriminator is much better trained than the generator, the gradient
vanishes to zero and the generator is unable to get useful learning
information from it. To mitigate this convergence problem, Wasserstein
GAN(WGAN) was introduced by Martin Arjovsky
in~\cite{Arjovsky:arxiv'17}. It replaces the conventional
JS-Divergence with Wasserstein distance as the measurement of the
difference between real and generated data which solves the vanishing
gradient problem. In this work, we employ the Wasserstein distance in
the loss functions to help the convergence. It also mitigates the
collapse mode problem to some extent and ensures the diversity of the
generated data.

\subsection{EM-GAN Architecture}
\label{sec:emganarchitecture}
\begin{figure}[!htb]
\vspace{-0.1in}
\centering
\includegraphics[width=0.8\columnwidth]{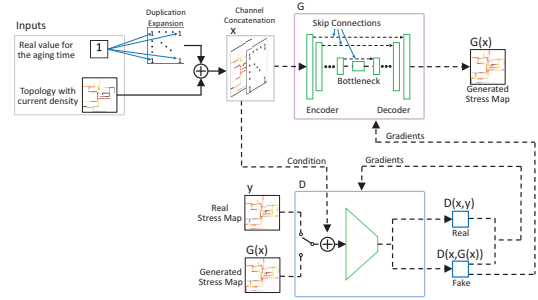}
\vspace{-0.05in}
\caption{EM-GAN framework for stress estimation}
\vspace{-0.1in}
\label{fig:emganframework}
\end{figure}

To implicitly learn the distribution of the current density image and
map it to the corresponding real-like EM stress image, we use a CGAN
as backbone for our model shown in Fig.~\ref{fig:emganframework}. The
generator takes the current density image $\boldsymbol{img_{cur}} \in
\mathbb{R}^{256 \times 256 \times 1}$ and the aging years
$\boldsymbol{t} \in \mathbb{R}$ as input. $\boldsymbol{t}$ is expanded
into $\mathbb{R}^{256 \times 256 \times 1}$ by channel-wise
duplication, such that $\boldsymbol{img_{cur}}$ and $\boldsymbol{t}$
can be concatenated depth-wise. The resulting input $\mathbf{x}$ given
to the generator is a ${256 \times 256 \times 2}$ image with all
entries normalized as described in Section~\ref{sec:dataprep}. We
employ an encoder-decoder architecture as our generator which is
widely used in image-to-image applications. In such a network, the
input is downsampled through a series of convolutional layers until a
bottleneck layer, at which the latent features are extracted and then
reversely upsampled through transposed convolutional layers. All
information is supposed to be passed through this bottleneck layer,
which is not necessary, as much low-level information is shared
between the input and output. In our work, both input and output
images share the same topology of interconnect layout. To bypass the
bottleneck layer and shuttle such information directly across the
network, we add skip connections between the encoder and the decoder.
Such architecture greatly helps to improve the result accuracy which
is discussed in detail in
Section~\ref{sec:analysisoflossandskipconnections}.

The output of the generator is denoted as $\mathbf{G(x)}$. Either the
generated $\mathbf{G(x)}$ or the real EM stress image $\mathbf{y}$ is
fed into the discriminator $\mathbf{D}$ alternatively together with
its corresponding current density and aging time $\mathbf{x}$ as the
condition input. The output of the discriminator is denoted as
$\mathbf{D(G(x),x)}$ or $\mathbf{D(y,x)}$ depending on which EM stress
image, generated or real, was inputed. 

The key idea of the proposed EM-GAN model is to get the generator
learn the mapping method from the distribution of current density and
aging year to that of the EM stress image in the training set. Such
transformation is achieved by progressively training the generator
according to the gradients back propagated from the loss based on the
output of the discriminator. The generator and the discriminator are
trained simultaneously but based on separate loss functions. The
training goal of the discriminator is to minimize $\mathbf{D(G(x),x)}$
and maximize $\mathbf{D(y,x)}$, which means higher scores should be
given to the real EM stress images than the generated ones. This
training objective can be expressed mathematically as

\begin{equation}
\begin{aligned}
\max _{D} \{ &\mathbb{E}_{\mathbf{x}, \mathbf{y}}[D(\mathbf{y},
    \mathbf{x})]-\mathbb{E}_{\mathbf{x}}[D(G(\mathbf{x}), \mathbf{x})]-\\
    &\lambda_{g p} \mathbb{E}_{\hat{\mathbf{x}}}
    [(\|\nabla_{\hat{\mathbf{x}}} D(\hat{\mathbf{x}}, \mathbf{x}))\|_{2}-1)^{2}]\}
\label{eqn:loss_d}
\end{aligned}
\end{equation}

$\mathbb{E}_{\mathbf{x}, \mathbf{y}}$ and $\mathbb{E}_{\mathbf{x}}$
are the expectations over the distributions of $\mathbf{x}$ and
$\mathbf{y}$. The last term in \eqref{eqn:loss_d} is the gradient
penalty which is adopted from WGAN-GP~\cite{Arjovsky:arxiv'17}.
$\hat{\mathbf{x}}$ is interpolation between the generated EM stress
image and its ground truth. The hyperparameter $\lambda_{g p}$ is the
weight of the gradient penalty which maintains the 1-Lipschitz
continuity of the discriminator. 

On the contrary, the training objective for the generator is to
deceive the discriminator and get higher scores for its generated EM
tress images. As the generator has no influence on the scores of the
real images, term $\mathbf{D(y,x)}$ is discarded in its objective
function. We also add a L2-norm to the loss of the generator, as is
shown in \eqref{eqn:loss_g}, to further improve the objective function
according to~\cite{Isola:CVPT'17}. $\lambda_{L 2}$ controls the
strength of the L2-norm distance penalty on the loss of generator. 

\begin{equation}
\min _{G} \left \{ \mathbb{E}_{\mathbf{x}}[-D(G(\mathbf{x}),
\mathbf{x})]+\lambda_{L 2} \cdot \mathbb{E}_{\mathbf{x}, 
\mathbf{y}}[\|\mathbf{y}-G(\mathbf{x})\|_{2}] \right \}
\label{eqn:loss_g}
\end{equation}

\begin{figure}[!htb]
\centering
\begin{subfigure}{0.99\columnwidth}
\centering
\includegraphics[width=1\columnwidth]{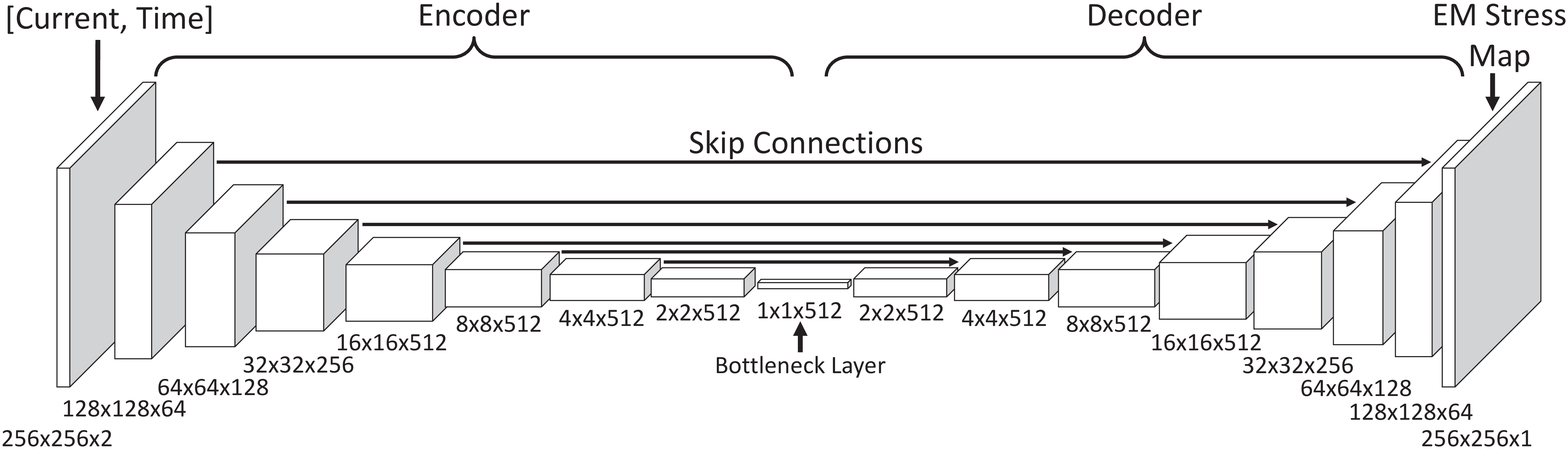}
\caption{}\label{fig:emgan_g}
\end{subfigure}
\vspace{+0.15in}
\begin{subfigure}{0.7\columnwidth}
\centering
\includegraphics[width=1\columnwidth]{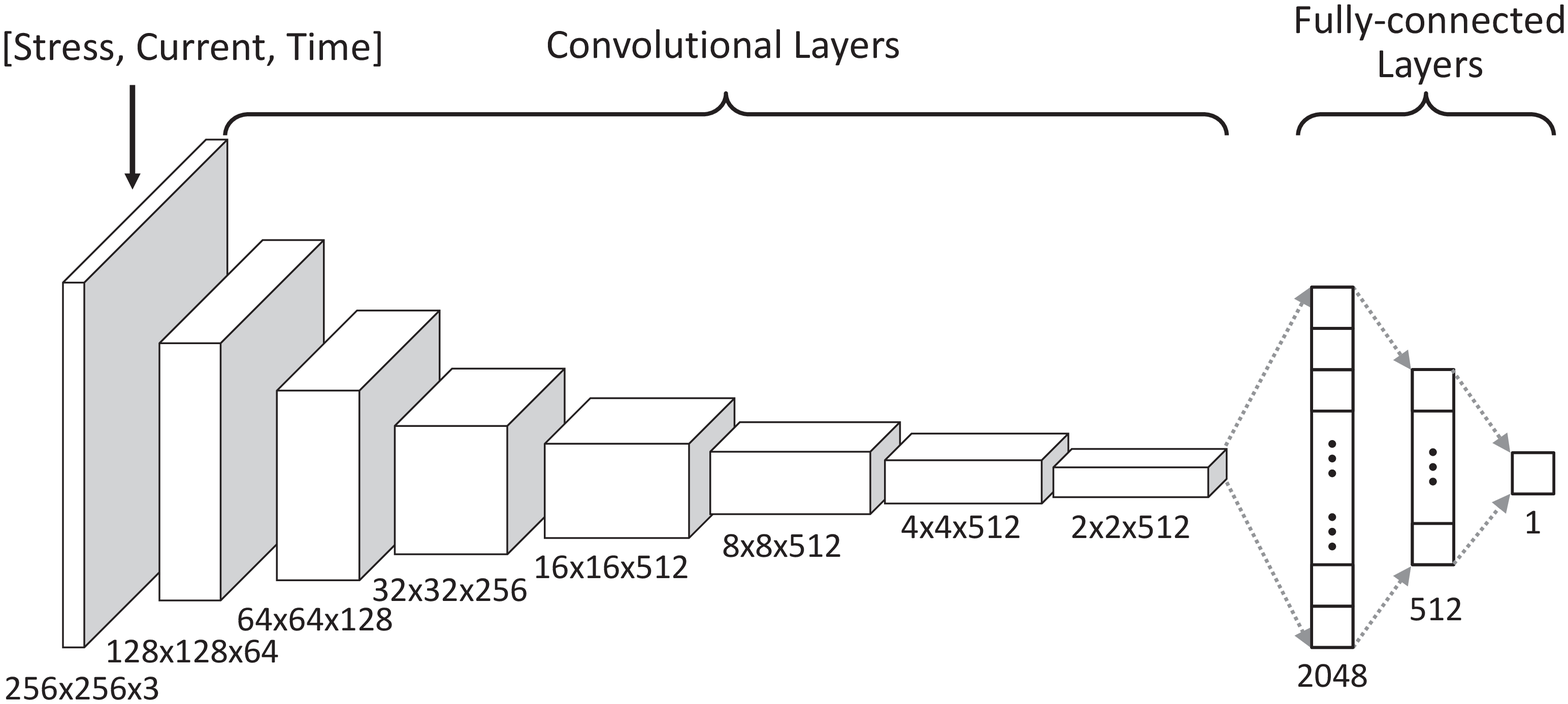}
\caption{}\label{fig:emgan_d}
\end{subfigure}
\caption{The architecture of the neural networks in the proposed EM-GAN: (a) generator (b) discriminator.}
\label{fig:arch_g_d}
\vspace{-0.15in}
\end{figure}

In both \eqref{eqn:loss_d} and \eqref{eqn:loss_g}, we use the
Wasserstein distance as the measurement of the difference between the
real and the generated EM stress image distribution to take advantage
of higher stability and convergence possibility. The detailed
architectures of the generator and the discriminator in our proposed
EM-GAN are illustrated in Fig.~\ref{fig:arch_g_d}.

\newcommand{\epoch}{15\xspace}

\section{Experimental results and discussions}
\label{sec:experimentalresultsanddiscussions}
In this section, we present the experimental results showing both the
accuracy and speed of our proposed EM-GAN model for time dependent EM
stress estimation.

All of our model is implemented in Python basing on TensorFlow(1.14.0)
library~\cite{Martin:OSDI'16} which is an open-source machine learning
platform. To train the GAN model, a dataset containing 25000 pairs of
(Current density image with topology and aging time, EM stress image)
samples is used. The samples were derived from 2500 different designs
of multi-segment interconnects. For each design, we collect the EM
stress maps simulated by COMSOL at 10 discrete aging time instants(1
to 10 years). Random selection of 15\% designs is set aside for
testing purpose and the remaining 85\% form the training set. During
the training phase, all samples are randomly permuted at the beginning
of every epoch.

We run the training for \epoch epochs on a Linux server with 2 Xeon
E5-2698v2 2.3GHz processors and Nvidia Titan X GPU. The cudnn library
is used to accelerate the training process on GPU. To employ
mini-batch stochastic gradient descent(SGD), we set the batch size to
8 and solve it with the RMSProp optimizer. The learning rate of the
optimizer is 0.0001, where the decay, momentum and epsilon parameters
are set to 0.9, 0 and 10$^{-10}$ respectively. The weight of the
L2-norm distance $\lambda_{L 2}$ is set to 100.

\subsection{Accuracy of EM Stress Map Estimation}
\label{sec:accuracyofemstressmapestimation}
Once the EM-GAN model is trained, the generator is preserved and serve
as the generative model. It can take any multi-segment interconnects
design as input and estimate the EM stress map at a given aging year.
To evaluate the estimation error against the ground truth, we employ
the root-mean-square error(RMSE) and the normalized RMSE(NRMSE) given
in \eqref{eqn:rmse} and \eqref{eqn:nrmse} as the Metrics. 

\begin{equation}
    RMSE = \sqrt{\frac{\sum_{(x,y) \in S}[\sigma(x,y)-\sigma'(x,y)]^2}{|S|}}
    \label{eqn:rmse}
\end{equation}

\begin{equation}
    NRMSE = \frac{RMSE}{\sigma_{max}-\sigma_{min}}
    \label{eqn:nrmse}
\end{equation}
where $\sigma$ and $\sigma'$ are the real and generated EM stress map
respectively. $S$ is the set containing all positions with an
interconnect and $|S|$ denotes the number of pixels in $S$.
$\sigma_{max}$ and $\sigma_{min}$ denote the maximum and minimum
stresses in the ground truth EM stress image.

We evaluate our trained EM-GAN model on the testing set which was set
aside during the training phase. The designs in the testing set were
randomly generated in the same way as the training set was produced.
The random generation process guarantees that there is no overlap of
either topology or current densities between these two datasets. It
means that all designs used for evaluation are unseen and absolutely
new to the model. This testing set makes our work more practically
meaningful, as in real applications, it is merely possible that the
design given to the model is identical to any design used for
training. Otherwise, it will make the model more of a memory system
leading to a low ability of generalization. 

\begin{figure*}[!htb]
\centering
\begin{subfigure}{0.9\textwidth}
\centering
\includegraphics[width=1\textwidth]{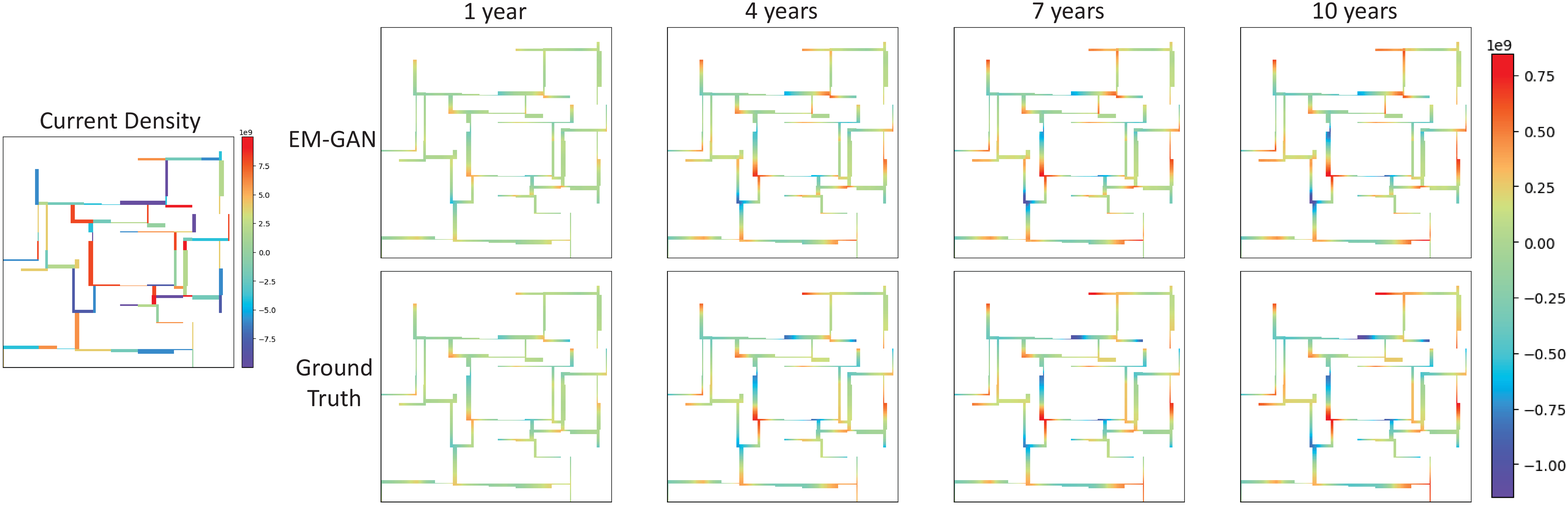}
\caption{}\label{fig:accuracy_design1}
\end{subfigure}
\vspace{+0.15in}
\begin{subfigure}{0.9\textwidth}
\centering
\includegraphics[width=1\textwidth]{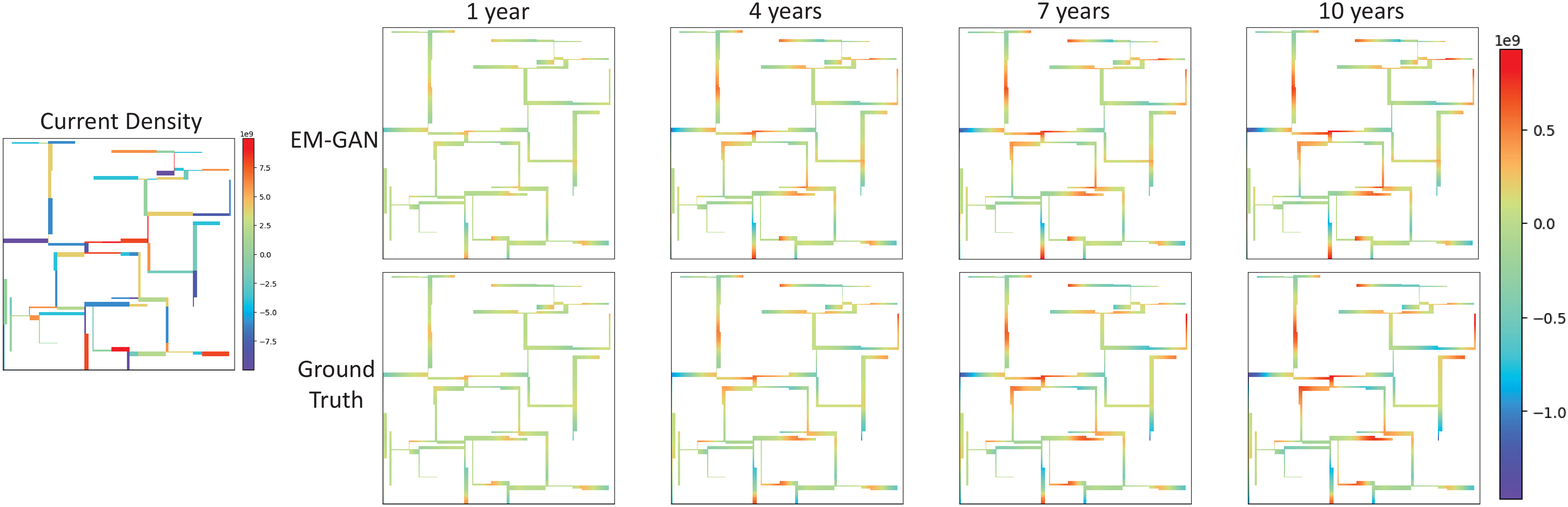}
\caption{}\label{fig:accuracy_design2}
\end{subfigure}
\vspace{-0.15in}
\caption{Comparing the ground truth EM stress distribution with EM-GAN generated
ones using two different designs.}
\label{fig:result_accuracy}
\vspace{-0.15in}
\end{figure*}

A total number of 375 different designs are tested and for each of
them, 10 EM stress images at 1 to 10 discrete aging years are
generated by our EM-GAN model. Comparing all 3750 generated EM stress
images with the ground truth, EM-GAN achieves an average estimation
RMSE of $0.13$ GPa and NRMSE of 6.6\%. Considering the large range the
values of typical EM stress vary in, usually several GPa, such
accuracy is beyond enough for EM failure assessment such as critical
wire identification. We randomly pick two testing designs and compare
the EM stress estimation at 1, 4, 7 and 10 aging years with the ground
truth COMSOL simulation results in Fig.~\ref{fig:result_accuracy}. The
unit of the current density is $A/m^{2}$ and EM stress is shown in
$Pa$.

\subsection{Speed of Inference}
\label{sec:speedofinference}
\begin{figure}[!htb]
\vspace{-0.1in}
\centering
\includegraphics[width=0.9\columnwidth]{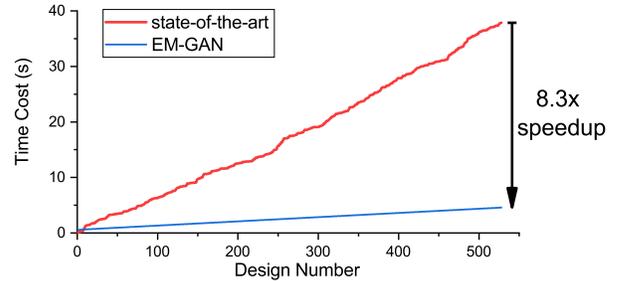}
\vspace{-0.05in}
\caption{Comparison of EM stress estimation speed between state-of-the-art and EM-GAN.}
\vspace{+0.05in}
\label{fig:result_speed}
\end{figure}

In what follows, we provide a comparison of speed between our EM-GAN
and the state-of-the-art work ~\cite{ChenTan:TVLSI'19} on EM stress
analysis. We set up the problem as a large multi-segment interconnects
design that can be divided into 528 smaller designs with same
dimensions of $256 \times 256$ $\upmu$m$^{2}$. We randomly pulled the
small designs from both training and testing set. The number of
branches in each design ranges from 5 to 79. Both our EM-GAN model and
the~\cite{ChenTan:TVLSI'19} method were tested to generate the EM
stress estimation at 10th aging year. The experiments were performed
on the same server and the accumulating time cost on all designs are
plotted in Fig.~\ref{fig:result_speed}.

The total time cost on the 528 small designs are 37.86s and 4.58s
for~\cite{ChenTan:TVLSI'19} and EM-GAN respectively. The EM-GAN
demonstrates $8.3 \times$ speedup over~\cite{ChenTan:TVLSI'19}.
For~\cite{ChenTan:TVLSI'19}, the time cost on the estimation of a
single design varies from 0.49s to 0.003s depending on the number of
branches in the design. For EM-GAN, there is no difference between
designs with different branches, and the inference speed is steadily
around 8ms per design throughout the experiment. The time cost of
EM-GAN is invariant to interconnect branches, which makes it much more
suitable for larger scale designs and leads to a better scalability.

\subsection{Analysis of Loss and Skip Connections}
\label{sec:analysisoflossandskipconnections}
As described in Section~\ref{sec:emganarchitecture}, our EM-GAN model
employs skip connections in the generator to bypass the bottleneck
layer in conveying the topology information from the encoder to the
decoder. We also applied L2-norm distance in the loss function of the
generator to enhance its performance. To analyze whether and how these
modifications are helpful to our application, we trained two other
modified models with similar structures as the proposed EM-GAN except
that one has no skip connection and the other discarded the L2-norm in
the objective function.

\begin{figure}[!htb]
\vspace{-0.1in}
\centering
\includegraphics[width=1.0\columnwidth]{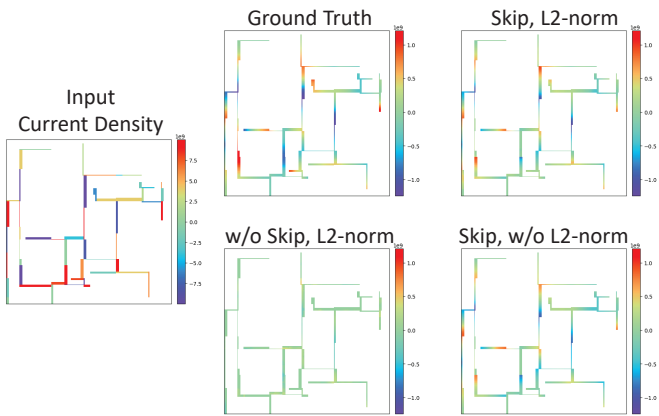}
\vspace{-0.05in}
\caption{Comparison of inference results between different models and the ground truth.}
\vspace{+0.05in}
\label{fig:modelcomparison}
\end{figure}

Both modified models are trained until convergence and tested against
the same training and testing set used by the EM-GAN. Both models
achieved worse errors on the testing set with NRMSE of 15.2\% for the
model without skip connection and 8.4\% for the one without L2-norm.
Also, our proposed EM-GAN demonstrates a smaller mean and standard
deviation in errors than the other two modified models, as shown in
Table~\ref{table:result_statistics}. In
Fig.~\ref{fig:modelcomparison}, we show the comparison between the
inference results generated by these models and the ground truth using
one randomly selected design from the testing set. Two models with
skip connections outperform the other one by a significant margin.
This can be accounted by the fact that the bottleneck layer handles
both structure and current information in conventional encoder-decoder
model. However, when skip connections are added, topology information
is directly passed from encoder to decoder and only current density
information is left to pass through the bottleneck. It greatly
increases the bandwidth of the information flow within the model and
helps increase the overall accuracy.

\renewcommand{\arraystretch}{1.3}
\begin{table}[h!]
\small
\centering
\caption{Statistics of NRMSE for EM-GAN and modified models on testing set.}
\begin{tabular}{| c | c | c | c |}
  \hline
  Metrics &  \makecell{EM-GAN\\(Skip, L2-norm)} &  \makecell{w/o
  Skip,\\L2-norm} &  \makecell{Skip,\\w/o L2-norm}\\
  \hline
  Mean               & 6.6\%  & 15.2\% & 8.4\%  \\
  \hline
  \makecell{Standard\\Deviation} & 1.2\%  & 2.1\%  & 2.1\%  \\
  \hline
  Max                & 12.9\% & 24.6\% & 18.4\% \\
  \hline
  Min                & 3.1\%  & 9.8\%  & 3.8\%  \\
\hline 
\end{tabular}
\label{table:result_statistics}
\end{table}

The model without L2-norm distance in its objective function
degenerates the NRMSE by a small margin from 6.6\% to 8.4\%. This can
be verified in Fig.~\ref{fig:modelcomparison} that both EM stress
images generated by the model with or without L2-norm are similar to
each other. However, the model with L2-norm has a much faster
converging speed in the training process and is always closer to the
ground truth than the one without L2-norm. It is a reasonable result
that the L2-norm helps the model as a prior knowledge. At the very
beginning of training process, both discriminator and generator are
not well trained and the discriminator is not able to provide useful
guidance to the generator. This is where L2-norm can complement the
discriminator and provide the generator with a meaningful learning
direction. In our experiment, adding the L2-norm accelerates the
convergence speed by $2\times$ and also leads to a better inference
accuracy.


\section{Conclusion}
\label{sec:concl}
In this paper, we have proposed a GAN-based fast transient hydrostatic
stress analysis for EM failure assessment for multi-segment
interconnects. In our approach, we treat this traditional numerical
PDE solving problem as time-varying 2D-image-to-image problem where
the input is the multi-segment interconnects topology with current
densities and the output is the EM stress distribution in those wire
segments at the given aging time. We randomly generated the training
set and trained the model with the COMSOL simulation results.
Different hyperparameters of GAN were studied and compared. After the
training process, the proposed EM-GAN model is tested against 375
unseen multi-segment interconnects designs and achieved high accuracy
with an average error of 6.6\%. It also showed $8.3 \times$ speedup
over recently proposed state of the art analytic based EM analysis
solver.

\bibliographystyle{unsrt}
\footnotesize
\bibliography{../../bib/softerror,../../bib/reinforcement,../../bib/stochastic,../../bib/simulation,../../bib/modeling,../../bib/reduction,../../bib/misc,../../bib/architecture,../../bib/mscad_pub,../../bib/thermal_power,../../bib/thermal,../../bib/reliability,../../bib/embedded,../../bib/machine_learning,../../bib/physical,../../bib/neural_network.bib}


\end{document}